# Class Imbalance Problem in Data Mining: Review

[1]Mr.Rushi Longadge, [2] Ms. Snehlata S. Dongre, [3]Dr. Latesh Malik

[1] Department of Computer Science and Engineering
G. H. Raisoni College of Engineering
Nagpur, India
rushi.longadge@gmail.com

[2] Department of Computer Science and Engineering
G. H. Raisoni College of Engineering
Nagpur, India
dongre.sneha@gmail.com

[3] Department of Computer Science and Engineering
G. H. Raisoni College of Engineering
Nagpur, India
latesh.malik@raisoni.net

## Abstract

In last few years there are major changes and evolution has been done on classification of data. As the application area of technology is increases the size of data also increases. Classification of data becomes difficult because of unbounded size and imbalance nature of data. Class imbalance problem become greatest issue in data mining. Imbalance problem occur where one of the two classes having more sample than other classes. The most of algorithm are more focusing on classification of major sample while ignoring or misclassifying minority sample. The minority samples are those that rarely occur but very important. There are different methods available for classification of imbalance data set which is divided into three main categories, the algorithmic approach, data-preprocessing approach and feature selection approach. Each of this technique has their own advantages and disadvantages. In this paper systematic study of each approach is define which gives the right direction for research in class imbalance problem.

## 1. Introduction

In many real time applications large amount of data is generated with skewed distribution. A data set said to be highly skewed if sample from one class is in higher number than other [1] [16]. In imbalance data set the class having more number of instances is called as major class while the one having relatively less number of instances are called as minor class [16]. Applications such as medical diagnosis prediction of rare but important disease is very important than regular treatment. Similar situations are observed in other areas, such as detecting fraud in banking operations, detecting network intrusions [10], managing risk and predicting failures of technical equipment.

In such situation most of the classifier are biased towards the major classes and hence show very poor classification rates on minor classes. It is also possible that classifier predicts everything as major class and ignores the minor class. various techniques have been proposed to solve the problems associated with class imbalance [9], which divided into three basic categories, the algorithmic approach, data-preprocessing and feature selection approach.

In data-preprocessing technique sampling is applied on data in which either new samples are added or existing samples are removed. Process of adding new sample in existing is known as over-sampling and process of removing a sample known as under-sampling. Second method for solving class imbalance problem is creating or modifying algorithm. The algorithms include the cost-sensitive method and recognition-based approaches, kernel-based learning, such as support vector machine (SVM) and radial basis function [16]. Applying an algorithm alone is not good idea because size of data and class imbalance ratio is high and hence a new technique i.e. the combination of sampling method with algorithm is used [12].

In classification, algorithm generally gives more important to correctly classify the majority class samples. In many applications miss-classifying a rare event can be result in more serious problem than common event [11]. For example in medical diagnosis in case of cancerous cell detection, misclassifying non-cancerous cells may leads to some additional clinical testing but misclassifying cancerous cells leads to very serious health risks. However in classification problems with imbalanced data, the minority class examples are more likely to be misclassified than the majority class examples, due to their design principles, most of the machine learning algorithms optimizes the overall classification accuracy which results in misclassification minority classes.

The paper is organized as follows: section 2 contains current approaches which gives basic techniques that used to solve the problem of imbalance dataset. Section 3 gives the review of related work that handle class imbalance problem. Section 4 gives comparative study of some algorithm and finally end with concluding conclusion in Section 5.

## 2. Current approaches

The literature survey suggests many algorithm and techniques that solve the problem of imbalance distribution of sample. These approaches are mainly dividing into three methods such as sampling, algorithms, and feature selection.

### 2.1 Sampling

Sampling techniques used to solve the problems with the distribution of a dataset, sampling techniques involve artificially re-sampling the data set, it also known as data preprocessing method. Sampling can be achieved by two ways, Under-sampling the majority class, oversampling the minority class, or by combining over and under-sampling techniques.
.
Under-sampling: The most important method in under-sampling is random under-sampling method which trying to balance the distribution of class by randomly removing majority class sample. Figure 1 show the random under-sampling method [4]. The problem with this method is loss of valuable information.

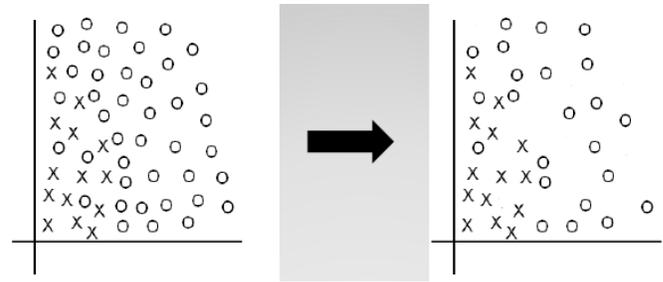

Fig.1.Randomly removes the majority sample.

Over-sampling:   Random Oversampling methods also help to achieve balance class distribution by replication minority class sample.

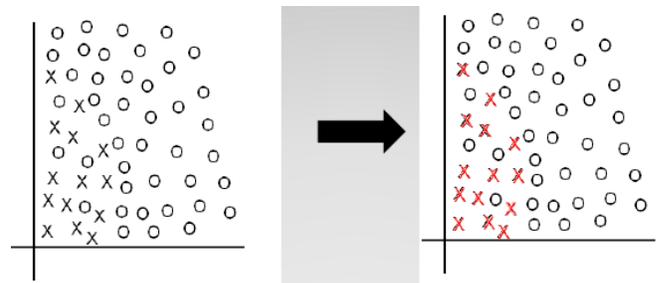

Fig.2. Replicate the minority class samples

There is no need to add extra information, it reuse the data [12]. This problem can be solving by generating new synthetic data of minority sample. SMOTE generates synthetic minority examples to over-sample the minority class. In this method learning process consume more time because original data set contain very small number of minority samples.

### 2.2 Algorithms

A several new algorithms have been created for solving the class imbalance problem. The goal of this approach is to optimize the performance of learning algorithm on unseen data. One-class learning methods recognized the sample belongs to that class and reject others. Under certain condition such as multi-dimensional data set one-class learning gives better performance than others [5]. Instated of changing class distribution applying cost in decision making is another away to improve the performance of classifier.   Cost-sensitive learning methods try to maximize a loss function associated with a data set. These learning methods are motivated by the finding that most real-world applications do not have uniform costs for misclassifications. The actual costs associated with each kind of error are unknown typically, so these methods need to determine the cost matrix based on the data and apply that to the learning stage. A closely related idea to cost-sensitive learners is shifting the bias of a machine to favor the minority class [8].

The goal of cost sensitive classification is to minimize the cost of misclassification, which can be realized by choosing the class with the minimum conditional risk. Table 1 gives the cost matrix which contains two classes $i \& j$. $\lambda_{ij}$ cost of misclassification. Diagonal element are Zero indicate that cost of correct classification has no cost. Another algorithmic approach for skewed distribution of data is modifying the classifier [8]. Kernel-based approach borrows the idea of support vector machines to map the imbalanced dataset into a higher dimension space. Then by combining with over-sampling technique or ensemble method, the classifier is supposedly to perform much better than learning from the original dataset.

Table 1: Cost matrix

| | | Prediction | |
|---|---|---|---|
| | | Class i | Class j |
| **True** | **Class i** | 0 | $\lambda_{ij}$ |
| | **Class j** | $\lambda_{ji}$ | 0 |

In terms of SVMs, several changes have been made to improve their class prediction accuracy and result suggests that SVM have ability to solve the problem of skewed vector without introducing noise [9]. Boosting methods can be combined with SVMs very effectively in the presence of imbalanced data [16].

### 2.3 Feature Selection

The goal of feature selection, in general, is to select a subset of j features that allows a classifier to reach optimal performance, where j is a user-defined parameter. For high-dimensional data sets, it uses filters that score each feature independently based on a rule. Feature selection is a key step for many machine learning algorithms, especially when the data is high-dimensional. Because the class imbalance problem is commonly accompanied the issue of high dimensionality of the data set, hence applying feature selection techniques is essential. Sampling techniques and algorithmic methods may not be enough to solve high dimensional class imbalance problems [5]. Feature selection as a general part of machine learning and data mining algorithms has been thoroughly researched, but its importance to resolving the class imbalance problem is a recent development with most research appearing in the previous several years [18]. In this time period, a number of researchers have conducted research on using feature selection to combat the class imbalance problem. Ertekin [17] studied the performance of feature selection metrics in classifying text data drawn from the Yahoo Web hierarchy. They applied nine different metrics to the data set and measured the power of the best features using the naïve Bayes classifier.

## 3. Related work

Data sampling has received much attention in data mining related to class imbalance problem. Data sampling tries to overcome imbalanced class distributions problem by adding samples to or removing sampling from the data set [2]. This method improves the classification accuracy of minority class but, because of infinite data streams and continuous concept drifting, this method cannot suitable for skewed data stream classification. Most existing imbalance learning techniques are only designed for two-class problem. Multiclass imbalance problem mostly solve by using class decomposition. AdaBoost.NC [1-4] is an ensemble learning algorithm that combines the strength of negative correlation learning and boosting method. This algorithm mainly used in multiclass imbalance data set. The results suggest that AdaBoost.NC combined with random oversampling can improve the prediction accuracy on the minority class without losing the overall performance compared to other existing class imbalance learning methods. Wang et al. proposed the classification algorithm for skewed data stream in [2], which shows that clustering sampling outperforms the traditional under-sampling, since clustering helps to reserve more useful information. However, the method cannot detect concept drifting. Chris [2] proposed that both sampling and ensemble technique are effective for improving the classification accuracy of skewed data streams. SVM-based one-class skewed data streams learning method was proposed in [6], which cannot work with concept drifting. Liu et al. [16] proposed one class data streams algorithm, which follows the single classifier approach and can be used to classify text streams. One of the most common data sampling techniques is Random Under-sampling. RUS simply removes examples from the majority class at random until a desired class distribution is achieved. RUSBoost is a new hybrid sampling and boosting algorithm for learning from skewed training data. RUSBoost provides a simpler and faster alternative to SMOTEBoost which is another algorithm that combines boosting and data sampling [2]. RUS decreases the time required to construct a model, which is benefit when creating an ensemble of models that is use in boosting.

RUSBoost presents a simpler, faster, and less complex than SMOTEBoost for learning from imbalanced data. SMOTEBoost combines a popular oversampling technique (SMOTE) with AdaBoost, resulting in a hybrid technique that increases the performance of its components. Infinitely imbalanced logistic regression [8] a recently developed classification technique that is named infinitely imbalanced logistic regression (IILR) acknowledges the problem of class imbalance in its formulation. Logistic regression (LR) is a commonly used approach for performing binary classification. It learns a set of parameters, {w0, and w1}, that maximizes the likelihood of the class labels for a given set of training data. When the number of data points belonging to one class far exceeds the number belonging to the other class, the standard LR approach can lead to poor classification performance. Cost-sensitive neural networks use sampling and threshold-moving method [8], this technique modify the distribution of training data such that cost of example calculated based on appearance of example. Threshold-moving tries to move the output threshold toward low cost classes such that examples with higher costs become harder to be misclassified. Threshold-moving is a good choice which is effective on all the data sets and can perform cost-sensitive learning even with seriously imbalanced data sets. Boosting SVM [20] in this algorithm, the classifier is produced from the current weight observation. For given instance, class prediction function which is design in terms of kernel function K. Algorithm calculates the G-mean of classifier by applying different weight and generates new set of classifier. The weight is calculated in iteration of boosting algorithm. Finally, G-mean is used for prediction of good classifier from ensemble classifier. SVM boosting algorithm is still unable to handle the issue of imbalance distribution of data. For online classification of data streams with imbalanced class distribution, Lei [7] proposed an incremental LPSVM termed DCIL-IncLPSVM that has robust learning performance under class imbalance. Linear Proximal support vector machines [LPSVM], like decision trees, classic SVM, etc. are originally not design to handle drifting data streams that exhibit high and varying degrees of class imbalance. Learning from class imbalance data stream, incremental learning algorithm is desirable to pose a capability for dynamic class imbalance learning (DCIL), i.e. learning from data to adjust itself adaptively to handle varied class imbalances. Lei [7] proposes a new incremental learning of wLPSVM for DCIL, where non-stationary imbalanced stream data mining problem is formalized as learning from data chunks of imbalanced class ratio, which are becoming available in an incremental manner. The proposed DCIL-IncLPSVM updates its weights and LPSVM simultaneously whenever a chunk of data is presented or removed.

## 4. Discussion

Analysis drawn from comparative study of each of the algorithm is shown in following table.

Table 2: Comparative Study

| Sr. No | Algorithm | Advantages | Disadvantages |
|---|---|---|---|
| 1 | AdaBoost.NC[1] | Improve prediction accuracy of minority | Ignore overall performance of classifier |
| 2 | RUSBoost [2] | Simple, faster and less complex than SMOTE Boost algorithm | Unable to solve Multiclass imbalance problem |
| 3 | Infinitely imbalanced logistic regression [6] | Mostly used for binary classification | Performance is depends on number of outlier in data. |
| 4 | Linear Proximal support vector machines[7] | Handle dynamic class imbalance problem | No consideration for distribution of sample |
| 5 | BoostingSVM[20] | Improved the performance of SVM classifier for prediction minority sample | Ignore imbalance class distribution. |

Many areas are affected by class imbalance problems. The solution provided by many techniques in data mining is helpful but not enough. The consideration of which technique is best for handling a problem of data distribution is highly depends upon the nature of data used for experiment.

## 5. Conclusion

Practically, it is reported that data preprocessing provide better solution than other methods because it allow adding new information or deleting the redundant information, which helps to balance the data. Another method that helpful to solve the problem of class imbalance is boosting. Boosting is powerful ensemble learning algorithm that improved the performance of weak classifier. The algorithm such as RUSBoost, SMOTEBoost is an example of boosting algorithm. Feature selection method can also used for classification of imbalance data. The performance of a feature selection algorithm depends on the nature of the problem. Finally, this paper suggests that applying two or more technique i.e. hybrid approach gives better solution for class imbalance problem.

# References

[1] Shuo Wang, Member, and Xin Yao, "Multiclass Imbalance Problems: Analysis and Potential Solutions", IEEE Transactions On Systems, Man, And Cybernetics—Part B: Cybernetics, Vol. 42, No. 4, August 2012.

2] Chris Seiffert, Taghi M. Khoshgoftaar, Jason Van Hulse, and Amri Napolitano "RUSBoost: A Hybrid Approach to Alleviating Class Imbalance"IEEE Transactions On Systems, Man, And Cybernetics—Part A: Systems And Humans, Vol. 40, No. 1, January 2010.

3] Björn Waske, Sebastian van der Linden, Jón Atli Benediktsson, Andreas Rabe, and Patrick Hostert "Sensitivity of Support Vector Machines to Random Feature Selection in Classification of Hyper-spectral Data", IEEE Transactions On Geosciences And Remote Sensing, Vol. 48, No. 7, July 2010

4] Xinjian Guo, Yilong Yin1, Cailing Dong, Gongping Yang, Guangtong Zhou,"On the Class Imbalance Problem" Fourth International Conference on Natural Computation, 2008.

5] Mike Wasikowski, Member and Xue-wen Chen, "Combating the Small Sample Class Imbalance Problem Using Feature Selection", IEEE Transactions on Knowledge and Data Engineering, Vol. 22, No. 10, October 2010.

6] Rukshan Batuwita and Vasile Palade,"Fuzzy Support Vector Machines for Class imbalance Learning" IEEE Transactions On Fuzzy Systems, Vol. 18, No. 3, June 2010.

7] Lei Zhu, Shaoning Pang, Gang Chen, and Abdolhossein Sarrafzadeh, "Class Imbalance Robust Incremental LPSVM for Data Streams Learning" WCCI 2012 IEEE World Congress on Computational Intelligence June, 10-15,2012 - Australia.

8] David P. Williams, Member, Vincent Myers, and Miranda Schatten Silvious, "Mine Classification With Imbalanced Data", IEEE Geosciences And Remote Sensing Letters, Vol. 6, No. 3, July 2009.

9] Chris Seiffert, Taghi M. Khoshgoftaar, Jason Van Hulse, Amri Napolitano "A Comparative Study of Data Sampling and Cost Sensitive Learning" , IEEE International Conference on Data Mining Workshops. 15-19 Dec. 2008.

[10] Mikel Galar,Fransico, "A review on Ensembles for the class Imbalance Problem: Bagging,Boosting and Hybrid-Based Approaches" IEEE Transactions On Systems, Man, And Cybernetics—Part C: Application And Reviews, Vol.42,No.4 July 2012

[11] Yuchun Tang, Yan-Qing Zhang, Nitesh V. Chawla, , and Sven Krasser "Correspondence SVMs Modeling for Highly Imbalanced Classification" IEEE Transactions On Systems, Man, And Cybernetics—Part B: Cybernetics, Vol. 39, No. 1, February 2009

[12] Peng Liu, Lijun Cai, Yong Wang, Longbo Zhang "Classifying Skewed Data Streams Based on Reusing Data" International Conference on Computer Application and System Modeling (ICCASM 2010).

13] Zhi-Hua Zhou, Senior Member, and Xu-Ying Liu "Training Cost-Sensitive Neural Networks with Methods Addressing the Class imbalance Problem" IEEE Transactions On Knowledge And Data Engineering, Vol. 18, No. 1, January 2006

14] Qun Song Jun Zhang Qian Chi " Assistant Detection of Skewed Data Streams Classification in Cloud Security", IEEE Transaction 2010.

15] Nadeem Qazi,Kamran Raza, "Effect Of Feature Selection, Synthetic Minority Over-sampling (SMOTE) And Under-sampling On Class imbalance Classification", 14th International Conference on Modeling and Simulation- 2012.

16]. Nitesh V. Chawla, Nathalie Japkowicz, Aleksander Ko lcz "Special Issue on Learning from Imbalanced Data Sets" Volume 6, Issue 1 - Page 1-6.

17] S¸ eyda Ertekin1, Jian Huang, L´eon Bottou, C. Lee Giles "Active Learning in Imbalanced Data Classification"

18] Saumil Hukerikar, Ashwin Tumma, Akshay Nikam, Vahida Attar "SkewBoost: An Algorithm for Classifying Imbalanced Datasets" International Conference on Computer & Communication Technology (ICCCT)-2011.

19] Chris Seiffert, Taghi M. Khoshgoftaar, Jason Van Hulse, "Improving Learner Performance with Data Sampling and Boosting" 2008 20th IEEE International Conference on Tools with Artificial Intelligence.

20] Benjamin X. Wang and Nathalie Japkowicz "Boosting Support Vector Machines for Imbalanced Data Sets" Proceedings of the 20th International Conference on Machine Learning-2009.

**Mr.Rushi N. Longadge** received Bachelor of Engineering. degree in Information Technology from North Maharstra University, Jalgon, India in 2010. He is pursuing M.Tech degree in Computer science and Engineering from G.H. Raisoni College of Engineering, Nagpur, India. His research area is Data Mining and Machine Learning.

**S. S. Dongre** she received B.E degree in Computer science and Engineering from Pt. Ravishankar Shukla University, Raipur, India in 2007 and M.Tech degree in Computer Engineering from university of Pune, Pune ,India in 2010. She is currently working as Assistant professor in the Department of Computer science and Engineering at G.H. Raisoni College of Engineering, Nagpur, India. Number of publication is in IEEE and Journals. Her research is on Data Stream Mining, Machine Learning, Decision Support system, ANN and Embedded System. Her book has published on title Data Stream Mining: Classification and Application, LAP publication House, Germany, 2010. Ms. Snehlata S. Dongre is a member of IACIST, IEEE and ISTE Organization.

**Dr. Latesh Malik** She received B.E degree in Computer science and Engineering from Rajastan University, India in 1996 and M.Tech degree in Computer science and Engineering from Banasthali Vidyapith, India in 2001. She is currently working as Professor and head of Department in Department of Computer science and Engineering at G.H. Raisoni College of Engineering, Nagpur, India. Number of publication is in IEEE and Journals. Her research is on Document Processing, Soft Computing and Image Processing. She received Best teacher award and she is Gold medalist in M.Tech and B.E. she also received the grant from AICTE of Rs. 7.5 lacs under RPS. Dr. Latesh Malik is a member of CSI, IEEE, and ISTE Organization.